\documentclass[11pt,letterpaper]{article}
\usepackage{cogsys}
\usepackage[T1]{fontenc}
\usepackage{times}
\usepackage[pdftex]{graphicx} 
\usepackage{subcaption}
\usepackage{amsmath}
\usepackage{xcolor}
\usepackage{amsfonts}   
\usepackage{booktabs}
\usepackage{hyperref}
\usepackage{natbib}
\makeatletter
\def\@makefnmark{\hbox{\@textsuperscript{\normalfont\@thefnmark}}}
\makeatother

\newcommand{\starfootnotemark}{%
  \begingroup
    \renewcommand{\thefootnote}{\fnsymbol{footnote}}%
    \footnotemark[1]%
  \endgroup
}

\cogsysheading{12}{2025}{}{08/2025}{09/2025}

\ShortHeadings{Hierarchical Semantic Retrieval}
              {A.\ Gupta\textsuperscript{*}, K.\ Singaravadivelan\textsuperscript{*}, Z.\ Wang}

\begin{document} 

\title{Hierarchical Semantic Retrieval with Cobweb}
 
\author{Anant Gupta\starfootnotemark}{agupta886@gatech.edu}
\author{Karthik Singaravadivelan\starfootnotemark}{ksingara3@gatech.edu}
\author{Zekun Wang}{zekun@gatech.edu}

\address{College of Computing, Georgia Institute of Technology, 
         Atlanta, GA 30332 USA}

\begingroup
  \renewcommand{\thefootnote}{\fnsymbol{footnote}}
  \footnotetext[1]{Equal contribution.}
\endgroup

\vskip 0.2in

\begin{abstract}
Neural document retrieval often treats a corpus as a flat cloud of vectors scored at a single granularity, leaving corpus structure underused and explanations opaque. We use Cobweb--a hierarchy-aware framework--to organize sentence embeddings into a prototype tree and rank documents via coarse-to-fine traversal. Internal nodes act as concept prototypes, providing multi-granular relevance signals and a transparent rationale through retrieval paths. We instantiate two inference approaches: a generalized best-first search and a lightweight path-sum ranker. We evaluate our approaches on MS MARCO and QQP with encoder (e.g., BERT/T5) and decoder (GPT-2) representations. Our results show that our retrieval approaches match the dot product search on strong encoder embeddings while remaining robust when kNN degrades: with GPT-2 vectors, dot product performance collapses whereas our approaches still retrieve relevant results. Overall, our experiments suggest that Cobweb provides competitive effectiveness, improved robustness to embedding quality, scalability, and interpretable retrieval via hierarchical prototypes.
\end{abstract}

\section{Introduction}
\let\thefootnote\relax
\footnotetext{Code available at \url{https://github.com/Teachable-AI-Lab/cobweb-language-embedding}}
Humans naturally organize knowledge into hierarchies and reason with prototypes: representative exemplars that capture the central tendency of a concept while allowing graded membership and basic-level advantages in categorization \citep{rosch1975family,rosch1978principles}. 
These cognitive principles suggest that effective information access should benefit from \emph{both} hierarchical structure (topics $\rightarrow$ subtopics) and prototype-based reasoning (representative exemplars that summarize clusters).
Classic conceptual clustering methods, most notably \textsc{Cobweb} \citep{fisher1987cobweb}, formalized these ideas by incrementally constructing a classification tree whose internal nodes summarize data with concept-level statistics and whose leaves capture finer distinctions. 

In large-scale document retrieval, modern systems have shifted from lexical matching to neural embedding methods.
BM25 is a sparse lexical ranking function that scores query–document matches using term frequency, inverse document frequency, and document-length normalization \citep{robertson2009bm25}.
More recent neural approaches learn contextual representations that capture semantics with the Transformer architecture \citep{vaswani2023attentionneed}: cross-encoders (e.g., BERT re-rankers) model full query–document interactions for high accuracy \citep{nogueira2019passage}, and dual-encoder dense retrievers map both into a shared vector space for scalable nearest-neighbor search \citep{karpukhin2020dpr,reimers2019sentencebert}.
Late-interaction models such as ColBERT further improve effectiveness vs. efficiency trade-offs by matching token-level representations \citep{khattab2020colbert}.

Despite their popularity, most neural retrievers treat the corpus as a flat cloud of points in Euclidean space, with relevance computed by a single similarity at a single granularity. 
However, flat retrievers underutilized the corpus’ inherent topic structure; hierarchical indexes such as HNSW \citep{malkov2018hnsw} enable coarse-to-fine retrieval by searching from upper to lower layers.
Prototype-based reasoning can provide concept-level anchors that make ranked results and re-ranking decisions interpretable without sacrificing performance \citep{anand2022explainableIRsurvey}. 

Our work bridges the gaps between recent neural document retrieval approaches and human-like interpretability with Cobweb, a prototype- and hierarchy-aware retrieval framework that learns a hierarchical ``database'' over document embeddings.
Cobweb incrementally organizes the corpus into a tree whose internal nodes store intermediate prototypes that summarize their descendants.
At query time, retrieval proceeds coarse-to-fine: the query is matched to high-level prototypes and then refined down the tree to leaf documents. 
Central to this process is a hierarchical retrieval mechanism that aggregates prototype similarities along the query’s traversal path, yielding multi-granular relevance signals and an auditable explanation.

Our contributions are the following. 
1. We introduce hierarchical retrieval approaches that compose coarse-to-fine prototype similarities along a learned tree using Cobweb, generalizing flat vector-space matching to multi-level relevance and enabling efficient search using learned sentence embeddings from neural models.
2. We show how learning intermediate prototypes improves interpretability by exposing concept-level rationales.
3. We ground our approach on the MS MARCO~\citep{bajaj2016msmarco} and QQP~\citep{wang2018glue} datasets, using encoder-only, decoder-only, and encoder–decoder Transformer embeddings. Our hierarchical retrieval methods match or improve upon inner-product-based dense-retrieval baselines across embedding architectures, scale efficiently with both time complexity and data size, and yield human-interpretable prototype paths.

\section{Related Works}

\subsection{Language Representations}
Language representations aim to encode linguistic meaning in a form usable by learning algorithms. 
Early distributional semantics approaches were centered around sparse and symbolic bag-of-words with TF–IDF weighting and $n$-gram language models. 
These approaches captures term frequency and short-range co-occurrence but ignoring broader context and treating words as independent types.

Recent embedding-based approaches address these limitations by mapping tokens to dense vectors whose geometry reflects meaning. 
Work on Word2Vec \citep{mikolov2013efficient} and GloVe \citep{pennington2014glove} showed that words can be embedded in a continuous space where proximity correlates with semantic similarity. 
The introduction of Transformers \citep{vaswani2023attentionneed} enabled contextualized embeddings via self-attention, substantially improving performance across natural language processing tasks such as sentiment analysis and machine translation.


Encoder-only Transformer models such as BERT \citep{devlin2019bertpretrainingdeepbidirectional} and RoBERTa \citep{liu2019roberta} produce bidirectional representations that are widely used for retrieval. Decoder-only Transformer models such as GPT \citep{radford2019language} are primarily designed for generation but can provide usable embeddings from internal activations. Encoder–decoder architectures, such as T5 \citep{raffel2023exploringlimitstransferlearning}, handle sequence-to-sequence tasks while producing high-quality contextual encodings. 
In this work, we compare and contrast our document retrieval approaches using the embeddings produced by these three types of Transformer models.

\subsection{Document Retrieval}

Document retrieval ranks a large corpus by relevance to a user query, typically using a retriever to produce top-$k$ candidates.
Classical sparse lexical methods such as BM25~\citep{robertson2009bm25} rank documents by counting query-term matches, weighting rare terms more, and adjusting for repeated terms and document length, favoring exact lexical overlap.
Dense neural retrieval encodes queries and documents as continuous vectors and retrieves by nearest-neighbor search under inner product or cosine similarity \citep{karpukhin2020dpr}; late-interaction models such as ColBERT \citep{khattab2020colbert} retain token-level granularity while preserving efficient dot-product search. 

At scale, nearest-neighbor search relies on specialized indexes which are optimized for this dot product objective deeply ingrained into most embeddings models.
FAISS \citep{johnson2017billionscalesimilaritysearchgpus} provides exact (e.g., \texttt{IndexFlatIP}) and approximate indexes for dot-product search, and has been an industry standard owing to its robust implementation and matrix-operable parallelization. 
Graph-based, approximate nearest-neighbor structures like HNSW \citep{malkov2018hnsw} enable coarse-to-fine exploration. 
These measures are effective because they utilize an inner-product calculation (usually the dot-product).
However, these systems also largely treat the corpus as a flat set of vectors and rely on a single-step geometric similarity. 
In contrast, our approach organizes documents into a semantic hierarchy with interpretable prototypes, yielding multi-granular relevance signals and transparent retrieval paths that complement flat nearest-neighbor search and reranking.

\subsection{Hierarchical Clustering with Cobweb}

Hierarchical clustering organizes data into nested groups, such as a tree or a directed acyclic graph, that support analysis at multiple levels of granularity. 
Early agglomerative methods \citep{sneath1957application,ward1963hierarchical} construct such structures by iteratively merging items under a distance metric and linkage criterion, without learned prototypes or online adaptation. 

Cobweb \citep{fisher1987cobweb} instead performs incremental hierarchical clustering on concepts: it maintains a probabilistic taxonomy whose nodes summarize attribute distributions, and inserts each instance by maximizing category utility with create/merge/split/reorder operators. Because of Cobweb's nature of sorting top-down, it can be thought of as an unsupervised divisive strategy for concept formation, in contrast with agglomerative methods.
Extensions to Cobweb \citep{efficientinductionlanguagemodels2022, convolutionalcobweb2021, wang2025taxonomic} brought Cobweb forward as a substitute for neural applications and used its underlying architecture to imagine neural approaches, and recent studies show Cobweb’s robustness in vision and language tasks \citep{barari2024incrementalconceptformationvisual, barari2024avoiding, LIAN2025101371}. 
In this work, we repurpose Cobweb as a hierarchical database for document retrieval: internal nodes serve as interpretable prototypes that guide coarse-to-fine search and provide multi-granular relevance signals, offering a complementary alternative to flat nearest-neighbor retrieval.



\section{Methodology}

Our goal is to adapt Cobweb for large-scale semantic document retrieval, where documents and queries are represented not by discrete lexical features but by continuous, high-dimensional vectors. Modern neural retrieval methods rely on such \emph{latent semantic representations}, mapping text into a continuous space in which geometric proximity reflects semantic relatedness. This transformation is essential for moving beyond the lexical surface form: two documents discussing “car insurance” and “auto coverage” may share few or no exact terms, yet must be recognized as near-equivalent in meaning.  

We use the Cobweb/4V \citep{barari2024incrementalconceptformationvisual} algorithm for our experiments, which extends the original Cobweb incremental concept formation framework to support continuous features. 
However, applying Cobweb/4V in this setting introduces two challenges. First, prototypes within Cobweb/4V use diagonal Gaussian distributions, assuming that input features are conditionally independent.
This property is not guaranteed for neural embeddings optimized on the dot product, which often contain correlated dimensions. Second, the Cobweb/4V algorithm is designed to return a best-match concept that aggregates all retrieved nodes (including both intermediate and leaf ndoes) rather than returning all of the retrieved leaf nodes. 
Our methodology addresses both limitations: we preprocess embeddings with whitening to improve feature independence that can be captured by a diagonal variance structure, and we extend a new inference for Cobweb/4V to return multi-result ranking over its learned hierarchy. 

\subsection{Sentence Representations}
\label{sec:sentence_representations}
For Cobweb/4V to organize and search effectively, document vectors must capture high-level semantic relationships and yield interpretable prototypes that summarize coherent concepts. 
To achieve this, we use sentence embeddings derived from pre-trained Transformer-based language models that encode contextual meaning into dense vectors.  

Specifically, we treat a tokenized sentence $x_{1:T}=(x_1,\ldots,x_T)$ as input and obtain a vector from a pre-trained Transformer $f_\theta$ by pooling the representation of the final hidden states:
\(
z \;=\; g(f_{\theta}(x_{1:T}))\in\mathbb{R}^{d},
\)
where $f_\theta$ has fixed (pre-trained) parameters $\theta$, $g$ is a model-specific pooling function, and $z$ is the sentence embedding of dimension $d$.  

We evaluate three representative Transformer families: BERT \citep{devlin2019bertpretrainingdeepbidirectional} (encoder-only), GPT-2 \citep{radford2019language} (decoder-only), and T5 \citep{raffel2023exploringlimitstransferlearning} (encoder–decoder). While GPT-2 is primarily optimized for autoregressive generation rather than isotropic semantic embeddings, we include it to examine how generative-model representations perform in our retrieval framework and to serve as a contrasting baseline to embedding-oriented architectures. For BERT and T5, we test sentence-transformer variants \citep{reimers2019sentencebert}, which fine-tune encoders with a contrastive objective that better aligns embeddings with semantic similarity and perform better on similarity objectives.  
Further details of the model architectures and training objectives are provided in Appendix~\ref{app:sentence_embeddings}.

\subsection{Dimensional Independence of Representations}

Neural sentence embeddings, while effective for capturing semantics, often exhibit strong correlations across dimensions due to the way they are learned \citep{li2020sentenceembeddingspretrainedlanguageanisotropic}. 
However, Cobweb/4V assumes a diagonal covariance structure for the Gaussian parameters at each node.
To address this, we apply \emph{embedding whitening} techniques, including Principal Component Analysis (PCA) and Independent Component Analysis (ICA) \citep{yamagiwa2023discovering}. PCA not only helps decorrelate features but also allows optional dimensionality reduction, which can improve computational efficiency in large-scale retrieval settings. ICA further enhances isotropy—making the distribution more uniform and spherical—while reducing residual cross-dimensional correlations. 
By producing representations that can be better captured by a diagonal covariance structure, whitening ensures that the algorithm’s probabilistic category utility computations remain meaningful and stable in a dense-vector setting.

Although numerous other whitening methods exist, we focus on these relatively simple techniques to isolate and evaluate the general efficacy of whitening in our retrieval framework.

\subsection{Cobweb/4V Training}
\label{sec:cobweb_training}
Cobweb/4V incrementally constructs a hierarchy in which each internal node stores a \emph{prototype} as parameterized by Gaussian parameters $\mu$ and $\sigma^2$ that summarizes the instances (documents) beneath it. 
These prototypes are updated online as a new document inputs, and the Cobweb/4V selects among four possible operations at each step—create a new category, merge categories, split a category, or insert into an existing category using the information-theoretic \emph{category utility} (CU) metric \citep{corter1992explaining}:
\[
CU(c) = P(c) \,\big[ U(c_p) - U(c) \big],
\]
where \(U(c)\) denotes the entropy of concept \(c\) computed over its attribute distributions, and \(c_p\) is the parent of node \(c\). This measure balances category predictiveness and distinctiveness, guiding the construction of meaningful, discriminative prototypes.

After training, all raw document vectors extracted from language models are stored only at the leaf nodes, and each leaf corresponds to exactly one datapoint from the training set. Internal nodes do not store raw instances; instead, they maintain the probabilistic prototypes summarizing their descendants (the Gaussain parameters $\mu_c$ and $\sigma^2$ over sentence embeddings). This design ensures that retrieval can always access the original datapoints while still benefiting from the generalization properties of the internal prototypes.

The resulting hierarchy can be viewed as a hierarchical clustered database that encodes both fine-grained document instances and high-level semantic abstractions at internal nodes. We hypothesize that, when combined with appropriately whitened embeddings, this structure can improve prediction and retrieval accuracy by exploiting both local and global semantic organization.

\subsection{Hierarchical Prediction}
\label{sec:cobweb_prediction}
Once trained, the learned Cobweb/4V hierarchy functions as both a semantic index and a search structure over the embedding space. We devise and evaluate two prediction strategies.

Given a query vector \(x\in\mathbb{R}^d\), both retrieval methods rely on the \emph{collocation score} \citep{jones1983identifying}:
\(
s(c) = p(x \mid c) \, p(c \mid x),
\)
where \(c\) is a candidate concept node with Gaussian parameters $\mu_c$ and $\sigma^2_c$. This score reflects both how well the query matches the node’s prototype \(p(x \mid c)\) and how representative the node is of the query’s inferred category \(p(c \mid x)\) .
In this paper, we assume a uniform prior $p(c)$ over all concepts for a faster computation of $s(c)$. 
Specifically, applying Bayes’ rule:
\(
p(c \mid x) = \frac{p(x \mid c) \, p(c)}{\sum_{c'} p(c') \, p(x \mid c')},
\)
the collocation score simplifies to:
\[
s(c) = N \cdot p(x \mid c)^2,
\]
where \(N\) is a normalization constant independent of \(c\).  
Thus, ranking by \(s(c)\) is equivalent to ranking by \(p(x \mid c)\) alone.  
We exploit this simplification in both prediction methods described below.

\subsubsection{Generalized Best-First Search}
\label{sec:bfs}
We generalize the original Cobweb prediction methodology, which only predicts the single best element.  
Retrieval begins at the root and performs a greedy best-first search through the hierarchy using \(s(c)\) as the heuristic.
Instead of committing to a single greedy descent path, Cobweb/4V expands up to \(N_{\text{max}}\) nodes according to their collocation scores, enabling exploration of multiple promising branches. Let $\mathcal C^*$ be the collection of nodes as the result of expansion, and since documents are stored in the leaves, we rank the leaves in the order they are explored in $\mathcal{C}^*$ and return them.  


\subsubsection{Path Sum Prediction}
\label{sec:pathsum}
In this alternative approach, leaves are ranked by their cumulative path collocation scores. 
For a candidate leaf $\ell$, we define
\[
\mathrm{score}(\ell) = \sum_{i=1}^{|\text{path}(\ell)|} \log\!\big(s(c_i)\big),
\]
where $c_1, \dots, c_{|\text{path}(\ell)|}$ are the internal nodes on the path from the root to $\ell$. 
Unlike the generalized best-first search approach, which orders documents by the sequence in which nodes are expanded, 
path-sum prediction ranks documents directly by their path scores.

\section{Experiment}

\subsection{Experimental Setup}

\paragraph{Datasets.}
We evaluated our approach and baselines on two retrieval tasks: MS MARCO \citep{bajaj2016msmarco} and QQP \citep{wang2018glue}.
MS MARCO is a large-scale web search corpus built from real Bing queries and web content. We use its passage collection as our document pool, which contains 8.8M passages and 7k testing queries. We use it in an ad hoc retrieval setup: given a natural-language query (e.g., ``how to clean a laptop keyboard''), rank passages from the document pool by relevance and return the top-$k$ results.
QQP is a paraphrase identification dataset of paired Quora questions. It contains 364k training pairs, 40k validation pairs, and 391k test pairs. We recast it as a retrieval task where, for a given query question (e.g., ``How can I become a better cook?''), the objective is to retrieve its paraphrase as the relevant ``document'' from a candidate pool.

\paragraph{Metrics.}
We report Recall@$x$, Mean Reciprocal Rank (MRR@$x$), and Normalized Discounted Cumulative Gain (nDCG@$x$) for $x\in\{5,10\}$ on both MS MARCO and QQP, following prior work~\citep{khattab2020colbert}. Mean Reciprocal Rank (MRR) computes the reciprocal of the rank of the first relevant document for each query, averaged across queries. Higher values indicate that at least one relevant document appears closer to the top of the ranking. Additionally, Normalized Discounted Cumulative Gain (nDCG) accounts for both document relevance and position, rewarding highly relevant documents that appear earlier. 
Scores are normalized so that $1$ indicates a perfect ranking, making nDCG particularly useful when relevance is graded.




\subsection{Compared Approaches and Implementation}

\paragraph{Baselines.}
To effectively test the dot product, we use Facebook AI Similarity Search (FAISS) as our baseline, a library which optimizes exact-nearest-neighbors search by dot product through matrix-operable parallelization. Specifically, we employ the flat index (\texttt{IndexFlatIP}), which performs an exact dot-product search. This serves as a high-accuracy baseline for efficient similarity search over vector embeddings.

\paragraph{Cobweb/4V Variants.}
For each transformer model, we train two Cobweb/4V models as described in Section~\ref{sec:cobweb_training}. 
One model uses raw transformer embeddings, the other uses PCA and ICA whitened embeddings.
We evaluate two prediction methods on these hierarchies: \textbf{Cobweb-BFS} that performs the best-first Search method described in Section \ref{sec:bfs} and \textbf{Cobweb-PathSum} that performs path sum prediction as described in Section \ref{sec:pathsum}.

\paragraph{Shared Implementation Details.}
In all experiments, queries and candidate documents are embedded using the language models described in Section~\ref{sec:sentence_representations}, and retrieval is performed in the resulting embedding space. For Cobweb variants, embeddings are whitened prior to tree construction, with an explained variance threshold of $0.96$ used throughout. For dataset construction, we select a corpus of $n$ documents and $q$ queries such that each query has at least one corresponding answer document in the corpus. 
\paragraph{Architecture Comparison.}
To compare retrieval effectiveness across model architectures, we evaluate all approaches on a controlled setting with a fixed corpus of 10k documents and 1k queries for both QQP and MS~MARCO. This setup allows us to isolate the impact of different embedding models (RoBERTa, GPT-2, T5) and retrieval strategies (Dot Product, Cobweb-BFS, Cobweb-PathSum) without confounding effects from corpus size. The goal of this experiment is to assess how our retrieval approaches perform relative to a strong flat index baseline across a variety of embedding geometries.

\paragraph{Scaling Experiments.}
To study robustness at larger scales, we fix the embedding model architecture and vary the corpus and query sizes. By separating the architecture comparison from scaling analysis, we first evaluate our retrieval approaches' ability to leverage sentence embeddings in a fair, fixed-size regime, and then assess how well those gains hold when moving to larger, more realistic datasets.

\section{Results and Discussion}
\subsection{Comparing Embedding Models} 
To analyze the performance of our approaches on document retrieval tasks using different sentence embeddings, we report three retrieval metrics, recall, MRR, and nDCG, at $k=5$ and $k=10$ on two retrieval datasets: QQP and MS MARCO in Table~\ref{tab:qqp} and Table~\ref{msmarco} respectively. 
\begin{table}[th]
\centering
\footnotesize
\caption{QQP retrieval metrics at @k=5 (top) and @k=10 (bottom). All values are percentages. $^\dagger$: No whitening is applied.}
\setlength{\tabcolsep}{6pt}
\begin{tabular}{lccccccccc}
\toprule
& \multicolumn{3}{c}{RoBERTa} & \multicolumn{3}{c}{T5} & \multicolumn{3}{c}{GPT-2} \\
\cmidrule(lr){2-4}\cmidrule(lr){5-7}\cmidrule(lr){8-10}
\textbf{Method} & Recall & MRR & nDCG & Recall & MRR & nDCG & Recall & MRR & nDCG \\
\midrule
\multicolumn{10}{l}{\textit{@k=5}}\\
Dot Product$^\dagger$ (FAISS)             & \textbf{86.80} & \textbf{74.23} & \textbf{77.07} & \textbf{84.90} & \textbf{73.75} & \textbf{76.18} & 0.00 & 0.00 & 0.00 \\
Cobweb-BFS$^\dagger$ & 11.10 &  8.11 &  8.93 & 11.10 &  7.85 &  8.74 & 18.00 & 13.86 & 14.86 \\
Cobweb-PathSum$^\dagger$ & 56.30 & 44.06 & 46.91 & 53.40 & 40.97 & 44.05 & 20.40 & 14.78 & 15.90 \\
Dot Product (FAISS) & 86.70 & 74.40 & 77.43 & 84.70 & 73.66 & 76.04 & 19.90 & 10.61 & 12.85 \\
Cobweb-BFS                & \textbf{85.90} & \textbf{73.62} & \textbf{76.37} & \textbf{84.60} & 72.83 & 75.38 & \textbf{35.10} & \textbf{26.38} & \textbf{28.18} \\
Cobweb-PathSum            & 84.90 & 73.22 & 75.57 & 84.00 & \textbf{72.97} & \textbf{75.42} & 33.60 & 24.82 & 26.58\\
\midrule
\multicolumn{10}{l}{\textit{@k=10}}\\
Dot Product$^\dagger$ (FAISS)                   & \textbf{91.30} & \textbf{74.84} & \textbf{77.18} & \textbf{90.00} & \textbf{74.47} & \textbf{76.76} & 0.20 & 0.02 & 0.08 \\
Cobweb-BFS$^\dagger$ & 14.60 &  8.54 & 10.01 & 13.80 &  8.23 &  9.45 & 21.70 & 14.34 & 15.80\\
Cobweb-PathSum$^\dagger$ & 69.60 & 45.81 & 50.54 & 63.10 & 42.27 & 46.27 & 24.80 & 15.37 & 16.81 \\
Dot Product (FAISS) & 91.40 & 75.02 & 77.69 & 90.70 & 74.47 & 76.98 & 27.80 & 11.61 & 14.75 \\
Cobweb-BFS                & \textbf{91.00} & \textbf{74.30} & \textbf{76.75} & \textbf{89.90} & 73.57 & 75.95 &  \textbf{42.00} & \textbf{27.30} & \textbf{29.74}\\ 
Cobweb-PathSum            & 90.60 & 74.01 & 76.27 & 89.30 & \textbf{73.71} & \textbf{76.09} & 40.80 & 25.74 & 28.25\\ 
\bottomrule
\label{tab:qqp}
\end{tabular}
\end{table}

\begin{table}[th]
\centering
\footnotesize
\caption{MS MARCO retrieval metrics at @k=5 (top) and @k=10 (bottom). All values are percentages. $^\dagger$: No whitening is applied.}
\setlength{\tabcolsep}{6pt}
\begin{tabular}{lccccccccc}
\toprule
& \multicolumn{3}{c}{RoBERTa} & \multicolumn{3}{c}{T5} & \multicolumn{3}{c}{GPT-2} \\
\cmidrule(lr){2-4}\cmidrule(lr){5-7}\cmidrule(lr){8-10}
\textbf{Method} & Recall & MRR & nDCG & Recall & MRR & nDCG & Recall & MRR & nDCG \\
\midrule
\multicolumn{10}{l}{\textit{@k=5}}\\
Dot Product (FAISS)$^\dagger$                         & \textbf{89.70} & \textbf{62.15} & \textbf{66.53} & \textbf{92.50} & \textbf{64.35} & \textbf{68.66} & 0.00 & 0.00 & 0.00 \\
Cobweb-BFS$^\dagger$      & 13.70 &  8.79 &  9.70 & 18.40 & 12.07 & 13.17 &  \textbf{0.60} &  \textbf{0.32} &  \textbf{0.40} \\
Cobweb-PathSum$^\dagger$ & 72.80 & 47.45 & 52.29 & 73.70 & 47.82 & 53.13 & \textbf{0.60} & 0.32 & 0.39 \\
Dot Product (FAISS) & 88.90 & 60.98 & 66.17 & 88.30 & 57.97 & 63.33 & 0.40 & 0.20 & 0.25 \\
Cobweb-BFS                     & \textbf{87.80} & \textbf{59.58} & \textbf{64.32} & \textbf{88.90} & \textbf{61.32} & \textbf{65.62} & 0.40 & 0.18 & 0.25 \\
Cobweb-PathSum                 & 80.70 & 55.04 & 59.22 & 81.00 & 54.88 & 59.30 &  \textbf{0.60} &  0.18 &  0.29 \\
\midrule
\multicolumn{10}{l}{\textit{@k=10}}\\
Dot Product (FAISS)$^\dagger$                      & \textbf{98.90} & \textbf{63.48} & \textbf{65.93} & \textbf{99.30} & \textbf{65.31} & \textbf{66.99} & 0.00 & 0.00 & 0.00 \\
Cobweb-BFS$^\dagger$      & 17.50 &  9.29 & 10.47 & 24.20 & 12.85 & 14.50 &  0.70 &  0.33 &  0.40 \\
Cobweb-PathSum$^\dagger$ & 85.60 & 49.21 & 53.60 & 86.50 & 49.63 & 54.43 & \textbf{1.00} & \textbf{0.38} & \textbf{0.52} \\
Dot Product (FAISS) & 98.30 & 62.36 & 65.73 & 98.60 & 59.41 & 62.84 & 0.50 & 0.21 & 0.26 \\
Cobweb-BFS                     & \textbf{97.00} & \textbf{60.91} & \textbf{63.67} & \textbf{98.60} & \textbf{62.73} & \textbf{65.20} & 0.90 & 0.25 & 0.41 \\
Cobweb-PathSum                 & 88.70 & 56.21 & 58.46 & 89.50 & 56.10 & 58.71 & \textbf{1.00} &  0.23 &  0.39 \\
\bottomrule
\label{msmarco}
\end{tabular}
\end{table}
\noindent Our results indicate that whitening on sentence embeddings is important for our retrieval approaches to form robust hierarchies with text embeddings.
Across three embedding models and two datasets, whitening improves Cobweb-BFS's retrieval metrics (e.g. recall$@k=5$: $85.90\%$ vs. $11.10\%$ on QQP with RoBERTa embedding and recall$@k=10\%$: $98.60\%$ vs. $24.20\%$ on MS MARCO with T5 embedding), suggesting the whitening techniques we used effectively removes feature dependencies from language model's embeddings.
On the other hand, dot product methods do not benefit from a whitened embedding extracted from a RoBERTa or a T5 model. This is because the embeddings from these models are already optimized for the dot product similarity. As a result, whitening primarily normalizes the variance of each dimension and has little effect on the embedding geometry in the vector space.   
However, whitening benefits dot product methods on anisotropic embeddings from GPT-2 \citep{ethayarajh-2019-contextual} by removing dominant correlations and rescaling each direction to unit variance, yielding a representation more suitable for dot-product similarity.

\begin{figure}[ht]
    \centering
    \includegraphics[width=1.0\textwidth]{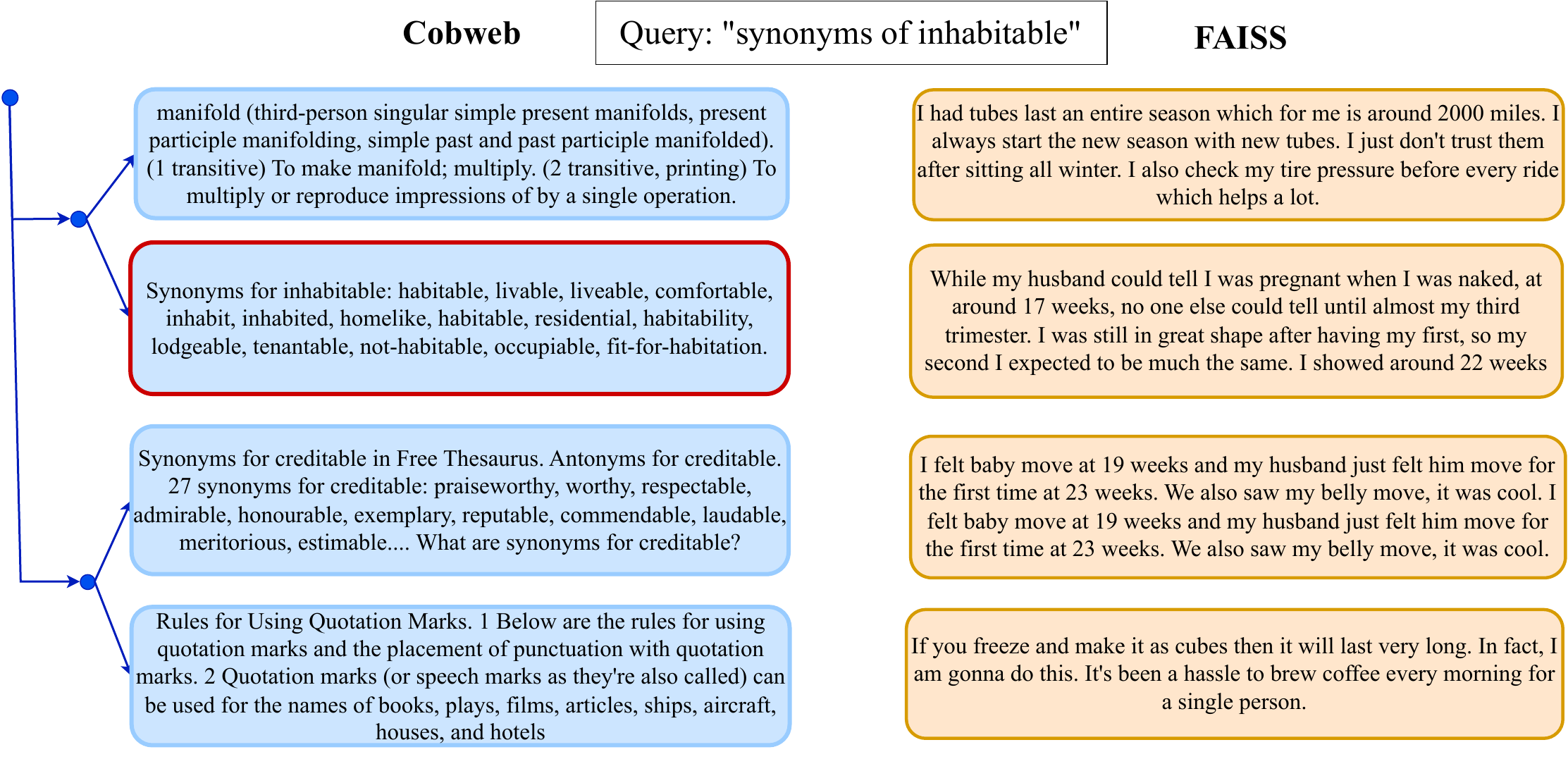}
    \caption{Examples of learned sub-hierarchies of whitened GPT-2 embeddings from the MS-MARCO dataset showcasing how the Cobweb-BFS metric appropriately retrieves relevant documents on the query "synonyms of inhabitable" while the dot product fails to retrieve relevant documents. The correct document is highlighted in red.}
    \label{fig:gptcasestudy1}
\end{figure}

Furthermore, our results show that our path sum prediction (Cobweb-PathSum) achieves performance comparable to the generalized best-first-search (Cobweb-BFS), with Cobweb-PathSum outperforming Cobweb-BFS on MRR and nDCG on both datasets using T5 embeddings. 
However, the path sum prediction offers an empirically lower time complexity than the generalized best-first-search, which leads faster runtime in practice. We refer Section~\ref{sec:time} for additional complexity analysis.
Across different datasets, our approaches match the dot product performance with RoBERTa and T5 embeddings, as these embeddings are optimized to well-align with similarity metrics, while offering a hierarchical document database. 
Interestingly, the dot product fails on both datasets using GPT-2 embeddings while our approaches robustly build hierarchies and retrieve.
We argue that in addition to whitening, our approaches' multi-step aggregation at various intermediate levels further removes the anisotropic distribution. Figure \ref{fig:gptcasestudy1} shows that dot product search ends up retrieving a subset of irrelevant documents, while our approaches successfully retrieve the document given the query.
We note that all approaches perform poorly on MS MARCO when using GPT-2 embeddings, since the dataset contains free-form answers rather than paraphrased questions, which requires explicit modeling of query--answer similarity.

\subsection{Visualizations} 

\begin{figure}[ht]
    \centering
    \includegraphics[width=1.0\textwidth]{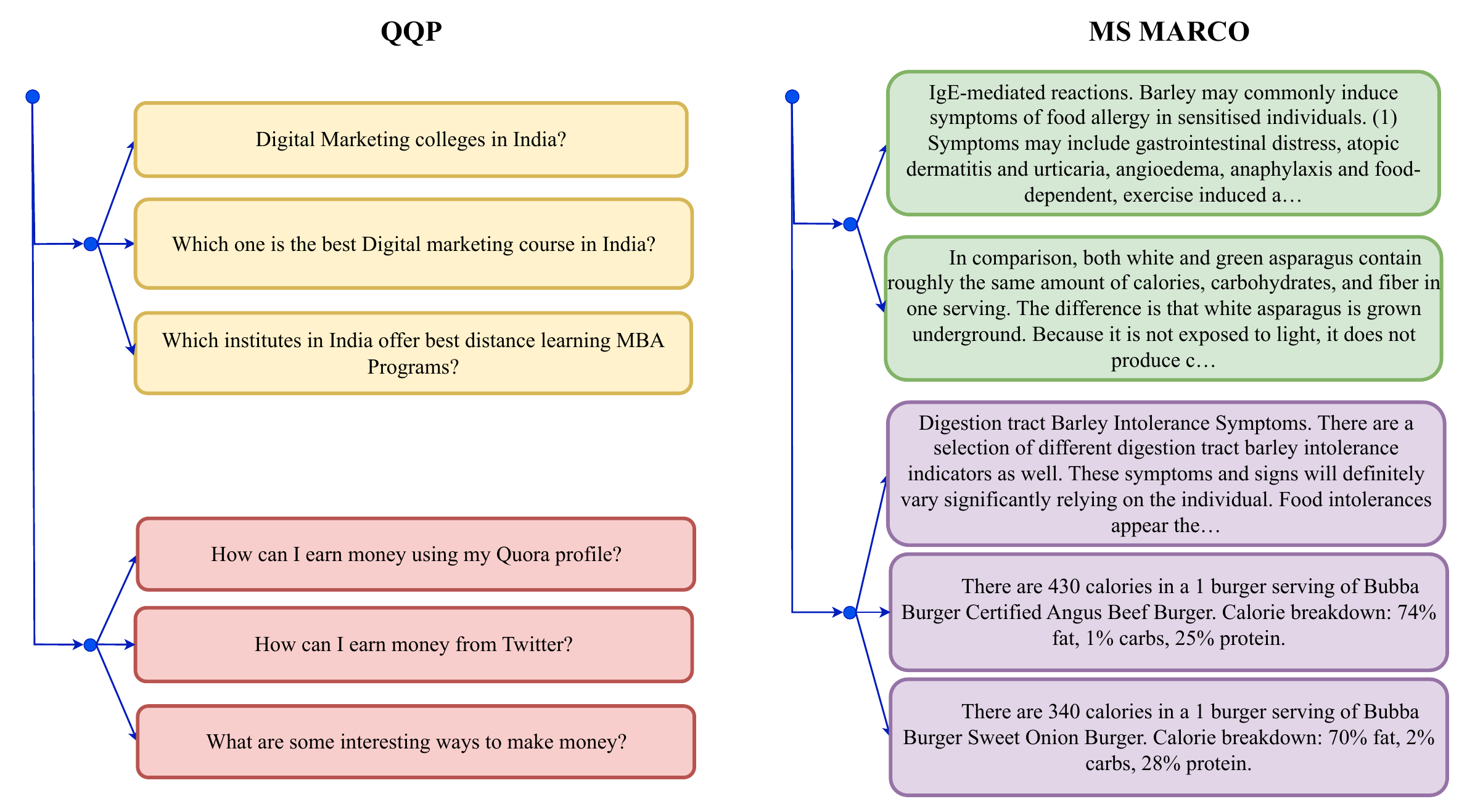}
    \caption{Examples of learned Cobweb/4V's sub-hierarchies on QQP and MS~MARCO using RoBERTa embeddings, with subtopics color‐coded by theme. Yellow: education in India, Red: earning money, Green: biochemistry, Purple: nutrition.}
    \label{fig:subtreeviz}
\end{figure}

Figure~\ref{fig:subtreeviz} shows examples of Cobweb/4V sub-hierarchies learned from QQP (left) and MS~MARCO (right) using RoBERTa embeddings. Subtopics are color-coded by theme (Yellow: education in India, Red: earning money, Green: biochemistry, Purple: nutrition). These visualizations indicate that the learned hierarchies capture multiple levels of semantic organization, with both fine-grained and broader prototypes emerging naturally from the data.

In the QQP example, questions about college educations in India (yellow) and questions about earning money through social media (red) appear as distinct sibling branches under a broader grandparent cluster.
Here, the concept of ``education'' is polysemous: it encompasses both the option of literal future education through a university system and the informal education of conducting a side hustle.
The grandparent prototype unites these interpretations under a generalized theme of “education and career,” while preserving their distinctions at the child level. 
Similarly, in the MS~MARCO example, passages about biochemical (green) and nutritional (purple) are grouped under a shared grandparent cluster corresponding to a broader “life sciences” theme. 
This higher-level prototype captures the conceptual connection between molecular biology and dietary science.

The presence of meaningful intermediate prototypes underscores that Cobweb/4V’s output is not a flat clustering, but a hierarchical semantic structure where each internal cluster summarizes and organizes its descendants.
This organization enables the hierarchy to represent relationships at multiple granularities: leaves capture specific document topics, immediate parents cluster closely related subtopics, and higher-level ancestors encode broader conceptual categories. 
Such multi-level organization suggests that the learned hierarchy serves as a semantic map of the corpus, providing interpretable structure that extends beyond mere proximity in the embedding space.

\subsection{Scalability and Efficiency Analysis}

We evaluate the scalability of our framework along two dimensions: \emph{accuracy at scale} and \emph{retrieval time complexity at scale}. The first measures how well retrieval quality is preserved as the number of indexed documents grows, while the second quantifies computational cost in practice.

\subsubsection{Corpus Scaling}
To evaluate the scalability of our framework, we conducted a corpus scaling experiment on the QQP dataset, comparing our approaches with a dot product (FAISS) baseline using RoBERTa embeddings. We measured the retrieval quality using Recall@10 and MRR@10 for different corpus sizes: 5k, 10k, 20k, and 40k documents. The results, as shown in Table~\ref{tab:scaling}, demonstrate that the retrieval performance of Cobweb, in both its BFS and PathSum variants, remains consistent with the dot product (FAISS) baseline across all corpus sizes.

As the corpus size increases from 5k to 40k, both FAISS and our retrieval approaches exhibit a gradual decrease in retrieval accuracy. This is a common phenomenon in information retrieval systems, as the number of potential matches for a query grows, making the task more challenging. For instance, the Recall@10 for FAISS drops from 96.60\% at 5k documents to 87.28\% at 40k documents. Similarly, the Recall@10 for Cobweb-BFS decreases from 96.60\% to 87.10\%, and Cobweb-PathSum from 96.00\% to 86.00\%. The MRR@10 metric shows a similar trend.

Crucially, the performance gap between FAISS and our retrieval approaches remains nearly constant. The difference in Recall@10 between FAISS and Cobweb-BFS, for example, is minimal across all scales, suggesting that Cobweb maintains its retrieval effectiveness even with a growing number of documents. While Cobweb-PathSum shows a slightly larger performance difference compared to FAISS, this difference does not widen as the corpus scales. These results indicate that our whitened framework is as robust to corpus size as the highly optimized FAISS baseline, preserving retrieval quality effectively at scale. We refer to Appendix \ref{app:scaling} for additional scaling results using T5 embeddings.
\begin{table}[th]
\centering
\footnotesize
\caption{QQP Scaling metrics @k=10 on RoBERTa Model Embeddings. All values are percentages. $\dagger$: No whitening is applied.}
\setlength{\tabcolsep}{6pt}
\begin{tabular}{lcccccccc}
\toprule
\textbf{Corpus Size} & \multicolumn{2}{c}{5k} & \multicolumn{2}{c}{10k} & \multicolumn{2}{c}{20k} & \multicolumn{2}{c}{40k} \\
 \cmidrule(lr){2-3}\cmidrule(lr){4-5}\cmidrule(lr){6-7}\cmidrule(lr){8-9}
\textbf{Method} & Recall & MRR & Recall & MRR & Recall & MRR & Recall & MRR \\
\hline
Dot Product$^\dagger$ (FAISS) & \textbf{96.60} & \textbf{80.22} & \textbf{91.30} & \textbf{74.84} & \textbf{88.48} & \textbf{69.93} & \textbf{87.28} & \textbf{68.35} \\
Cobweb-BFS & \textbf{96.60} & \textbf{80.03} & \textbf{91.00} & \textbf{74.30} & \textbf{87.92} & \textbf{69.28} & \textbf{87.10} & \textbf{67.80} \\
Cobweb-PathSum & 96.00 & 78.93 & 90.60 & 74.01 & 87.12 & 68.36 & 86.00 & 66.77 \\
\bottomrule
\label{tab:scaling}
\end{tabular}
\end{table}

\subsubsection{Time Analysis}
\label{sec:time}

Consider a database of $N$ documents, each represented by an embedding of dimension $D$. Let $k$ denote the number of top-ranked results to return, $d$ the depth of the Cobweb tree, and $N_0$ the number of nodes (internal and leaf) in the tree. We write $p(x \mid c)$ for the probability of a query $x$ under cluster $c$, assuming independent Gaussian features.

In practice, Cobweb/4V's hierarchies have branching factor $b>2$, so depth grows as $d = O(\log_b N)$ rather than linearly in $N$, and empirically we observe $N_0 \approx 1.5N$ across datasets. Moreover, retrieval tasks rarely require full sorting of all $N$ documents; since evaluation is limited to small cutoffs ($k \leq 1000$), partial ranking is sufficient.
Under these assumptions, we can compare the complexity of Dot product and our retrieval approaches as follows.

\paragraph{Dot Product (FAISS).}  
The similarity score is computed by calculating the dot product between the query and all document embeddings in $O(ND)$ time. Returning the top-$k$ results adds $O(N \log k)$, which is negligible compared to $ND$ when $D \gg \log k$. Thus, the effective complexity is $O(ND)$.

\paragraph{Weighted Path Sum retrieval (Cobweb tree).}
Each node score $\log p(c \mid x)$ is computed with simple $D$-dimensional vector operations: subtracting the prototype mean from the query, squaring the result, dividing by the node variance, and adding the log-variance term before reducing to a scalar. These element-wise additions, subtractions, and divisions are repeated across all nodes, giving $O(N_0D)$ complexity. A leaf’s score is then the sum of its ancestor scores along the path, with $O(d)$ accumulation per leaf and top-$k$ selection in $O(N\log k)$. With $N_0\!\approx\!1.5N$ and $d\!\ll\!D$, this reduces to $O(ND)$ overall.

\paragraph{Best-first search} 
Let $N_{\max}$ denote the maximum number of nodes expanded during search, as described in Section~\ref{sec:bfs}. At each step, the highest-scoring node is removed from a priority queue and expanded. Scoring a node requires $O(D)$ operations, while priority queue updates cost $O(\log N_{\max})$ each. Thus, visiting all $N_0$ nodes yields $O(N_0D + N_0\log N_{\max})$ complexity. Since $D \gg \log N_{\max}$, this reduces to $O(ND)$ in the worst case.  

In contrast, in the best case only a small fraction of nodes are expanded, following greedy retrieval. If search proceeds along a single high-scoring path, only $O(k \log N_0)$ nodes are visited, giving total complexity $O(kD \log N + N \log N_{\max})$. This makes best-first search potentially more efficient than exhaustive scoring when $k \ll N$.

\paragraph{Summary.}
Across all methods, the dominant cost arises from $D$-dimensional operations. Dot product search scales as $O(ND)$, weighted path sum retrieval as $O(ND)$, and best-first search as $O(ND)$ in the worst case with potential reductions to $O(N\log N_{\max})$ in favorable scenarios. This does not imply that they require similar absolute runtimes; rather, their scaling behavior differs only by multiplicative constants. These constants, influenced by memory layout, cache efficiency, and parallel matrix operations, can still result in substantial performance differences in practice. Thus, as shown in Table~\ref{tab:runtime-ms}, Cobweb-PathSum is parallelizable with matrix operations and achieves $20$--$100\times$ faster runtime than Cobweb-BFS.
\begin{table}[th]
\centering
\footnotesize
\caption{Average latency per query (in milliseconds) for different methods across corpus sizes on the MS MARCO dataset. $\dagger$: No whitening is applied.}
\setlength{\tabcolsep}{6pt}
\begin{tabular}{lccccc}
\toprule
\textbf{Method} & 1k & 5k & 7.5k & 10k & 20k \\
\hline
Dot Product$^\dagger$ (FAISS) & 0.40 & 1.27 & 1.31 & 3.03 & 5.91 \\
Cobweb-BFS & 164.45 & 724.52 & 1006.42 & 1524.63 & 3087.39 \\
Cobweb-PathSum & 1.69 & 11.54 & 14.37 & 27.25 & 52.63 \\
\bottomrule
\label{tab:runtime-ms}
\end{tabular}
\end{table}

\section{Conclusion and Future Works}
We introduced a hierarchy retrieval framework that adapts Cobweb/4V to dense textual neural embeddings, organizing documents into prototype trees for coarse-to-fine search. 
We proposed hierarchical retrieval approaches that compose prototype similarities at multiple levels of granularity.
Our approach achieved retrieval performance comparable to the dot product with strong encoder embeddings and remains robust in challenging settings, such as with anisotropic GPT-2 embeddings where the dot product fails. 
These hierarchies not only preserve accuracy at scale on both QQP and MS MARCO datasets but also provide interpretable multi-level relevance signals, bridging the gap between high-performance dense retrieval and human-like semantic organization.

This paper employs primitive techniques of whitening; however, future directions aim to employ whitening as a processing addition rather than a post-processing addition, such as through models that output whitened embeddings \citep{zhuo-etal-2023-whitenedcse}. Additional works could involve creating embeddings that are naturally optimized for Cobweb by integrating a differentiable Cobweb approximation into a model training process.

Methods of exact retrieval search the entire document space to return the top-k documents. In a large-scale exact-retrieval setting, many documents are irrelevant to the query, and so approximate-nearest-neighbors solutions have been introduced to calculate semantic similarities on a subset of the total database to improve efficiency, whether through hybrid approaches or greedy filtering \citep{malkov2018efficientrobustapproximatenearest, fu2016efannaextremelyfast, xu2025harmonyscalabledistributedvector}. Another future direction involves modifying the Cobweb metrics to approximate solutions by restricting our search to a specific set of leaf nodes or a specific sub-tree. 

Finally, the use of categorical utility and its inherent reliance on decorrelated dimensions is valuable in realizing the full value of an isotropic latent space, as analyzed by \citep{jung2023isotropicrepresentationimprovedense}. While isotropic embeddings are in theory superior because of their ability to explain more with fewer dimensions, the biggest barrier to their widespread use is finding metrics that inherently take advantage of independent dimensions \citep{yamagiwa2023discovering}. With our Cobweb metric, we open up the possibility of utilizing lower-dimensional, isotropic embedding spaces to describe a distribution that previously needed higher-dimensional descriptions.


 
\begin{acknowledgements} 
\noindent
We would like to thank Christopher J. Maclellan for discussions.
\end{acknowledgements}

\vspace{-0.2in}

{\parindent -10pt\leftskip 10pt\noindent
\bibliographystyle{cogsysapa}
\bibliography{main}

}

\appendix

\section{Sentence Embedding Architectures}
\label{app:sentence_embeddings}

We summarize the training objectives and pooling strategies for the three Transformer architectures considered.

\paragraph{Encoder-only (BERT).}
Trained with masked language modeling, predicting randomly masked tokens from bidirectional context:
\[
\max_{\theta}\;\sum_{t\in\mathcal{M}} \log p_{\theta}\!\big(x_t \mid x_{\setminus \mathcal{M}}\big),
\]
where $\mathcal{M}$ indexes masked positions. Sentence embeddings are formed by pooling the [CLS] token from the final encoder states.

\paragraph{Decoder-only (GPT-2).}
Models the joint distribution autoregressively with a left-to-right mask:
\[
\max_{\theta}\;\sum_{t=1}^{T} \log p_{\theta}\!\big(x_t \mid x_{<t}\big),
\]
where $x_{<t}$ is the prefix under a causal mask. We obtain sentence vectors by average pooling over final hidden states.

\paragraph{Encoder–decoder (T5).}
Uses a bidirectional encoder with cross-attention into a decoder, trained for conditional text generation:
\[
\max_{\theta}\;\sum_{t=1}^{T_y} \log p_{\theta}\!\big(y_t \mid y_{<t},\, x_{1:T_x}\big),
\]
where $x_{1:T_x}$ is the source sequence and $y_{1:T_y}$ the target. Sentence embeddings are taken from pooled encoder outputs.

\paragraph{Sentence-transformer variants.}
Fine-tune an encoder with a contrastive objective that pulls together semantically matched pairs and pushes apart in-batch negatives:
\[
\mathcal{L} = -\sum_i \log \frac{\exp(\mathrm{sim}(\hat z_i,\hat z_i^{+})/\tau)}{\sum_{j} \exp(\mathrm{sim}(\hat z_i,\hat z_j)/\tau)},
\]
where $\hat z$ is the \(\ell_2\)-normalized pooled embedding and $\mathrm{sim}$ is cosine similarity.

\section{Additional Visualizations}
\label{app:visualizations}

Below we denote some additional visualizations that showcase Cobweb's ability to cluster documents with semantic or syntactical similarity. 

\begin{figure}[h!]
    \centering
    \begin{subfigure}[t]{0.45\textwidth}
        \centering
        \includegraphics[width=\textwidth]{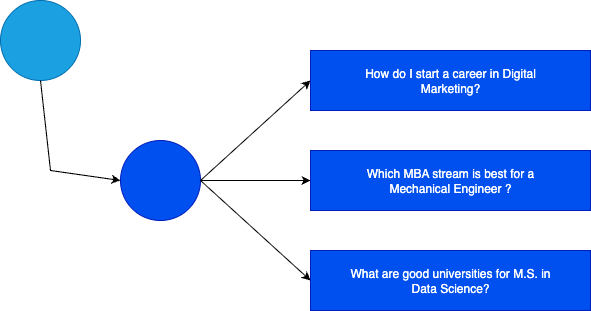}
        \caption{A subtree of whitened RoBERTa embeddings representing how three MS-MARCO paraphrases directed towards clarifying educational questions are all similarly represented.}
        \label{fig:education3}
    \end{subfigure}
    \hfill
    \begin{subfigure}[t]{0.45\textwidth}
        \centering
        \includegraphics[width=\textwidth]{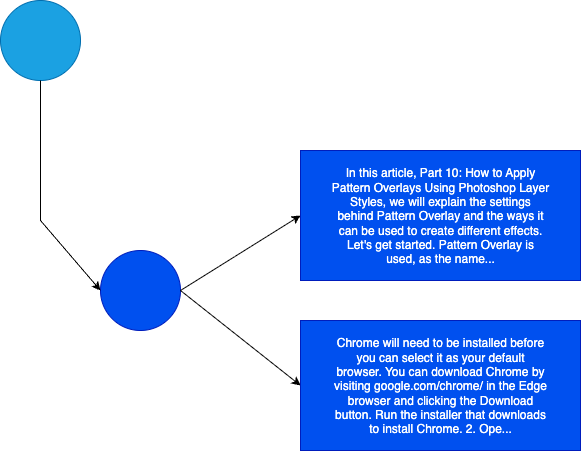}
        \caption{A subtree of whitened RoBERTa embeddings showcasing how two documents about setup instructions are clustered, despite referencing different subjects.}
        \label{fig:techinstructions2}
    \end{subfigure}

    \vskip\baselineskip
    \begin{subfigure}[t]{0.45\textwidth}
        \centering
        \includegraphics[width=\textwidth]{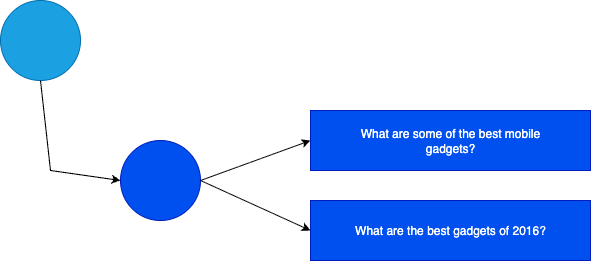}
        \caption{A subtree of whitened RoBERTa embeddings representing how two QQP queries about gadget rankings are closely intertwined.}
        \label{fig:gadget2}
    \end{subfigure}
    \hfill
    \begin{subfigure}[t]{0.45\textwidth}
        \centering
        \includegraphics[width=\textwidth]{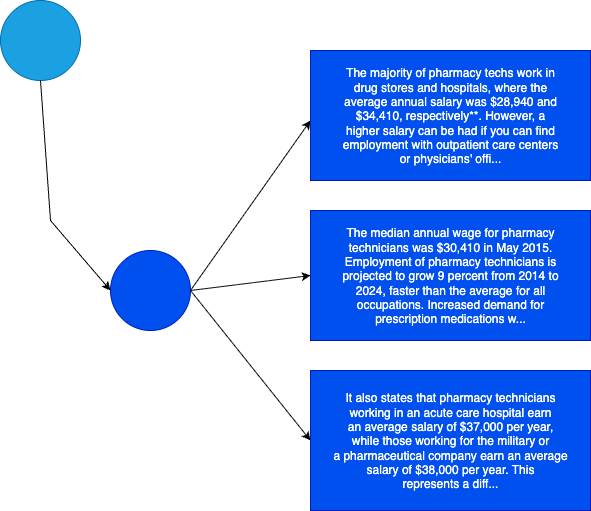}
        \caption{A subtree of whitened RoBERTa embeddings representing how three MS-MARCO paraphrases of the same document are correctly clustered under the same parent node.}
        \label{fig:pharmtech3}
    \end{subfigure}
    \label{fig:subtrees_part1}
\end{figure}

\begin{figure}[h!]
    \centering
    \begin{subfigure}[t]{0.45\textwidth}
        \centering
        \includegraphics[width=\textwidth]{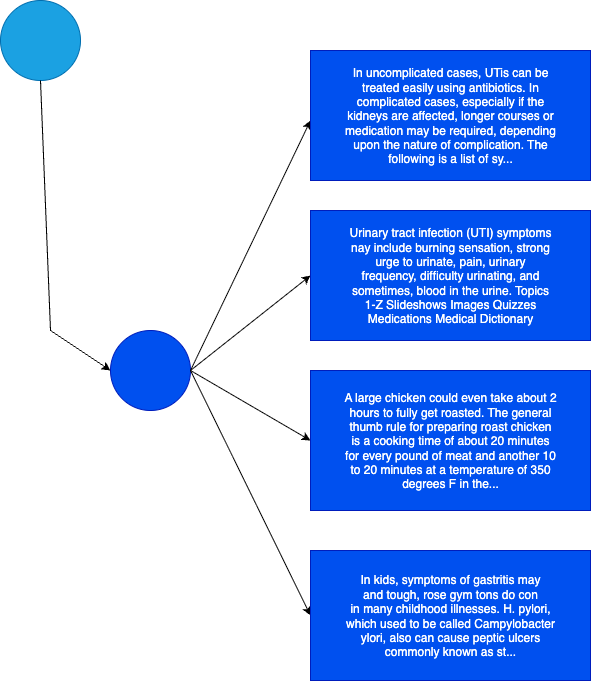}
        \caption{A subtree showing how the "burning sensation of UTIs" is semantically tied to the "roasting of chicken" – unintuitive behavior which results in optimal performance.}
        \label{fig:utiburningchicken4}
    \end{subfigure}
    \hfill
    \begin{subfigure}[t]{0.45\textwidth}
        \centering
        \includegraphics[width=\textwidth]{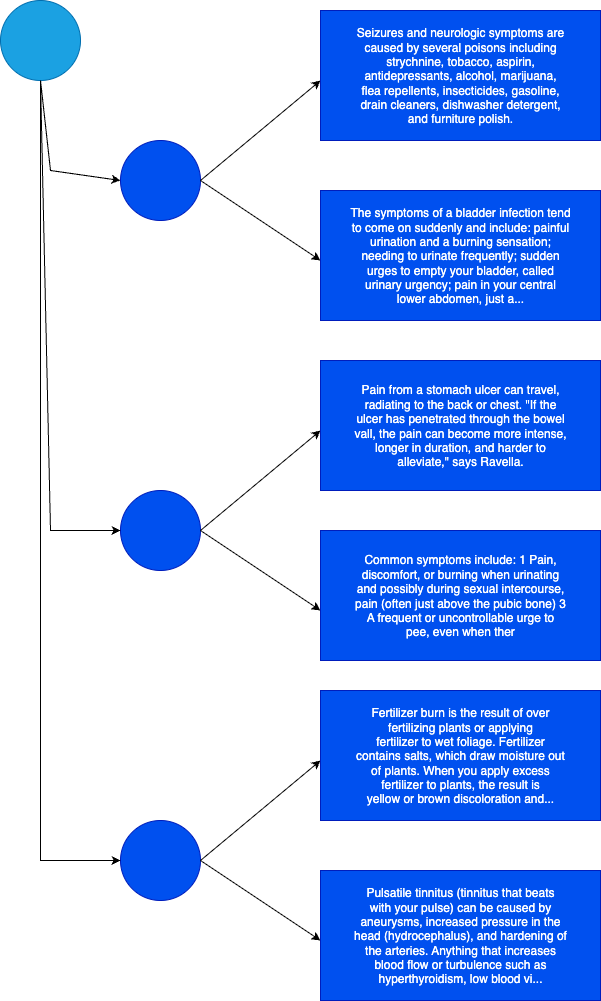}
        \caption{A subtree showing how six MS-MARCO paraphrases follow a "burning" and "pain" trend, despite discussing different topics.}
        \label{fig:burn6}
    \end{subfigure}
    \label{fig:subtrees_part2}
\end{figure}

\section{Additional Scaling Results}
\label{app:scaling}

Below we provide the scaling results on both MS MARCO (Table~\ref{tab:scaling_marco_app}) and QQP (Table~\ref{tab:scaling_qqp_app}) datasets on the gtr-t5-large sentence transformer. 

\begin{table}[h!]
\centering

\footnotesize
\caption{QQP retrieval metrics at @k=5 (top) and @k=10 (bottom) for T5 sentence transformer. All values are percentages. $\dagger$: No whitening is applied.}
\setlength{\tabcolsep}{6pt}
\begin{tabular}{lcccccccc}
\toprule
& \multicolumn{2}{c}{5k} & \multicolumn{2}{c}{10k} & \multicolumn{2}{c}{20k} & \multicolumn{2}{c}{40k} \\
\cmidrule(lr){2-3}\cmidrule(lr){4-5}\cmidrule(lr){6-7}\cmidrule(lr){8-9}
\textbf{Method} & Recall & MRR & Recall & MRR & Recall & MRR & Recall & MRR \\
\midrule
\multicolumn{9}{l}{\textit{@k=5}}\\
Dot Product$^\dagger$ (FAISS) & 87.60 & 77.26 & 84.90 & 73.75 & 80.88 & 68.96 & 79.60 & 67.23 \\
Cobweb-BFS & 86.80 & 76.68 & 84.60 & 72.83 & 80.64 & 68.44 & 78.95 & 66.55 \\
Cobweb-PathSum & 85.60 & 75.07 & 84.00 & 72.97 & 79.48 & 67.43 & 78.25 & 66.29 \\
\midrule
\multicolumn{9}{l}{\textit{@k=10}}\\
Dot Product$^\dagger$ (FAISS) & 93.20 & 78.02 & 90.00 & 74.47 & 86.64 & 69.73 & 85.88 & 68.05 \\
Cobweb-BFS & 93.60 & 77.60 & 89.90 & 73.57 & 85.84 & 69.15 & 85.65 & 67.43 \\
Cobweb-PathSum & 90.20 & 75.70 & 89.30 & 73.71 & 85.24 & 68.21 & 85.02 & 67.19 \\
\bottomrule
\label{tab:scaling_qqp_app}
\end{tabular}
\end{table}
\begin{table}[h!]
\centering

\footnotesize
\caption{MS MARCO retrieval metrics at @k=5 (top) and @k=10 (bottom) for T5 sentence transformer. All values are percentages. $\dagger$: No whitening is applied.}
\setlength{\tabcolsep}{6pt}
\begin{tabular}{lcccccccc}
\toprule
& \multicolumn{2}{c}{5k} & \multicolumn{2}{c}{10k} & \multicolumn{2}{c}{20k} & \multicolumn{2}{c}{40k} \\
\cmidrule(lr){2-3}\cmidrule(lr){4-5}\cmidrule(lr){6-7}\cmidrule(lr){8-9}
\textbf{Method} & Recall & MRR & Recall & MRR & Recall & MRR & Recall & MRR \\
\midrule
\multicolumn{9}{l}{\textit{@k=5}}\\
Dot Product$^\dagger$ (FAISS) & 92.40 & 62.40 & 92.50 & 64.35 & 92.81 & 66.15 & 91.80 & 65.02 \\
Cobweb-BFS & 88.20 & 58.55 & 88.90 & 61.32 & 90.47 & 62.91 & 89.38 & 61.62 \\
Cobweb-PathSum & 81.20 & 53.90 & 81.0 & 54.88 & 83.98 & 57.98 & 81.27 & 55.72 \\
\midrule
\multicolumn{9}{l}{\textit{@k=10}}\\
Dot Product$^\dagger$ (FAISS) & 99.80 & 63.45 & 99.30 & 65.31 & 98.95 & 67.04 & 98.22 & 65.95 \\
Cobweb-BFS & 98.40 & 60.05 & 98.60 & 62.73 & 97.41 & 63.9 & 97.20 & 62.74 \\
Cobweb-PathSum & 89.80 & 55.13 & 89.50 & 56.10 & 89.92 & 58.84 & 88.85 & 56.80 \\
\bottomrule
\label{tab:scaling_marco_app}
\end{tabular}
\end{table}

We additionally provide the runtime comparisons for different approaches across increasing corpus sizes. Table~\ref{tab:runtime-qqp} reports execution times (in seconds) on the QQP datasets (the MS MARCO version is in Table~\ref{tab:runtime-ms}).

\begin{table}[th]
\centering
\footnotesize
\caption{Average latency per query (in milliseconds) for different methods across corpus sizes on the QQP dataset. $\dagger$: No whitening is applied.}
\setlength{\tabcolsep}{6pt}
\begin{tabular}{lcccc}
\toprule
\textbf{Method} & 1k & 5k & 10k & 20k \\
\hline
Dot Product$^\dagger$ (FAISS) & 0.19 & 2.05 & 3.96 & 6.05 \\
Cobweb-BFS & 225.52 & 702.78 & 1418.06 & 2129.69 \\
Cobweb-PathSum & 1.41 & 15.03 & 53.05 & 139.81 \\
\bottomrule
\label{tab:runtime-qqp}
\end{tabular}
\end{table}

\end{document}